%% file: sample-sigconf.tex
\documentclass[
    sigconf = true,     
    review = false,      
    screen = true,      
    anonymous = false,   
]
{acmart}

\usepackage{booktabs} 
\usepackage{comment}
\usepackage{url}

\setcopyright{rightsretained}

\copyrightyear{2021}
\acmYear{2021}
\setcopyright{rightsretained} 
\acmConference[GECCO '21 Companion]{2021 Genetic and Evolutionary
Computation Conference Companion}{July 10--14, 2021}{Lille, France}
\acmBooktitle{2021 Genetic and Evolutionary Computation Conference
Companion (GECCO '21 Companion), July 10--14, 2021, Lille,
France}\acmDOI{10.1145/3449726.3459579}
\acmISBN{978-1-4503-8351-6/21/07}

\acmPrice{15.00}

\begin{document}
\title{OPTION: OPTImization Algorithm Benchmarking ONtology }

\author{Ana Kostovska}
\orcid{0000-0002-5983-7169}
\affiliation{%
  \institution{Jo\v{z}ef Stefan Institute \&}
  \institution{Jo\v{z}ef Stefan International Postgraduate School}
  \streetaddress{Jamova cesta 39}
  \city{Ljubljana}
   \country{Slovenia}
  \postcode{1000 }
}

\author{Diederick Vermetten}
\affiliation{
  \institution{Leiden Institute for Advanced Computer Science}
  \city{Leiden}
  \country{The Netherlands}
}

\author{Carola Doerr}
\orcid{0000-0002-5983-7169}
\affiliation{%
  \institution{Sorbonne Université, CNRS, LIP}
  \city{Paris}
   \country{France}
  \postcode{1000 }
}

\author{Sa\v{s}o D\v{z}eroski}
\orcid{0000-0003-2363-712X}
\affiliation{%
  \institution{Jo\v{z}ef Stefan Institute \&} 
  \institution{Jo\v{z}ef Stefan International Postgraduate School}
  \streetaddress{Jamova cesta 39}
  \city{Ljubljana}
  \country{Slovenia}
  \postcode{1000 }
}

\author{Pan\v{c}e Panov}
\orcid{0000-0002-7685-9140}
\affiliation{%
  \institution{Jo\v{z}ef Stefan Institute \&}  \institution{Jo\v{z}ef Stefan International Postgraduate School}
  \streetaddress{Jamova cesta 39}
  \city{Ljubljana}
   \country{Slovenia}
  \postcode{1000 }
}

\author{Tome Eftimov}
\orcid{}
\affiliation{%
  \institution{Jo\v{z}ef Stefan Institute}
  \streetaddress{Jamova cesta 39}
  \city{Ljubljana}
   \country{Slovenia}
  \postcode{1000 }
}

\renewcommand{\shortauthors}{A. Kostovska et al.}

\begin{abstract}
Many platforms for benchmarking optimization algorithms offer users the possibility of sharing their experimental data with the purpose of promoting reproducible and reusable research. However, different platforms use different data models and formats, which drastically inhibits identification of relevant data sets, their interpretation, and their interoperability.
Consequently, a semantically rich, ontology-based, machine-readable data model is highly desired. 

We report in this paper on the development of such an ontology, which we name OPTION (\textbf{OPTI}mization algorithm benchmarking \textbf{ON}tology). Our ontology provides the vocabulary needed for semantic annotation of the core entities involved in the benchmarking process, such as algorithms, problems, and evaluation measures.
It also provides means for automated data integration, improved interoperability, powerful querying capabilities and reasoning, thereby enriching the value of the benchmark data.
We demonstrate the utility of OPTION by annotating and querying a corpus of benchmark performance data from the BBOB workshop data -- a use case which can be easily extended to cover other benchmarking data collections.

\end{abstract}


\maketitle

\input{main_poster}

\bibliographystyle{ACM-Reference-Format}
\bibliography{sample-bibliography} 

\end{document}

%% file: main_poster.tex
\section{Introduction}

Evolutionary computation, and black-box optimization in general, is a rapidly growing field, which has seen tremendous progress in the last few decades, where many new algorithms are developed daily, making it impossible for researchers and practitioners in the field to stay up to date with all these developments and difficult to share their results with the research community.

There are unfortunately no common standards for \emph{which} performance data to record, nor \emph{how} to store it. Storing, sharing, and making the benchmark optimization data reusable is challenging, because of the existence of different data formats that are weakly compatible. There exist a large number of different benchmarking platforms for optimization, which have mostly been designed independently from each other. 

One well-established solution to the problem of managing such complex data pools is the use of ontologies. Ontologies are specifications of a shared conceptualization of data from distributed and heterogeneous systems and databases, and as such, they enable data interoperability, efficient data management and integration, and cross-database search. 

\section{The OPTION ontology}\label{sec:ontology}
In this paper, we present the \textbf{OPTimization Algorithm Benchmarking ONtology (OPTION)}. The ontology we propose was designed with the main goal of standardizing and formalizing knowledge from the domain of benchmarking optimization algorithms, where an emphasis was put on the representation of data from the benchmark performance space. OPTION offers a comprehensive description of the domain, covering the benchmarking process, as well as the core entities involved in the process, such as optimization algorithms, benchmark problems, evaluation measures, etc.  The OPTION ontology is publicly available on BioPortal\footnote{OPTION at BioPortal: \url{https://bioportal.bioontology.org/ontologies/OPTION-ONTOLOGY}} the largest repository of ontologies. We have also constructed a semantic annotation pipeline for annotating performance data, as well as a querying endpoint for querying the knowledge base. Finally, we have annotated benchmark performance data from 32 optimization algorithms and show how these annotations can be queried.

\section{Use case}

To demonstrate the utility of the OPTION ontology for semantic annotation, we consider the BBOB benchmark suite as a use case. We use a subset of BBOB data that includes the algorithms from the 2015-2017 competitions (three years). This gives us a total of 32 algorithms, which we semantically annotate using the OPTION ontology. All annotations were produced with the Apache Jena RDF library in the form of RDF graphs. For each of the 32 algorithms, a separate RDF graph was generated and uploaded to Apache Jena TDB2, a native triple store that efficiently stores RDF data.

The data in the ontology provides the full information on the used problem/dimension/instance/algorithm, as well as the corresponding performance data. In addition, data provenance information has been manually collected and linked to the performance data to trace its origin. The stored data provenance information includes the digital object identifier (DOI) of the paper, the paper's title, the authors' name, and the year of the publication.

\section{Querying the knowledge base}
After populating the knowledge base, we can use the SPARQL query language to ask queries. To enable this functionality, we have set up an Apache Jena Fuseki2 server and implemented two services. The query service provides an endpoint for handling SPARQL queries in a RESTful manner, while the upload service enables the upload of RDF data into the Apache Jena TDB2 triple store. 
\begin{figure}[b]
    \centering
    \includegraphics[width=\linewidth]{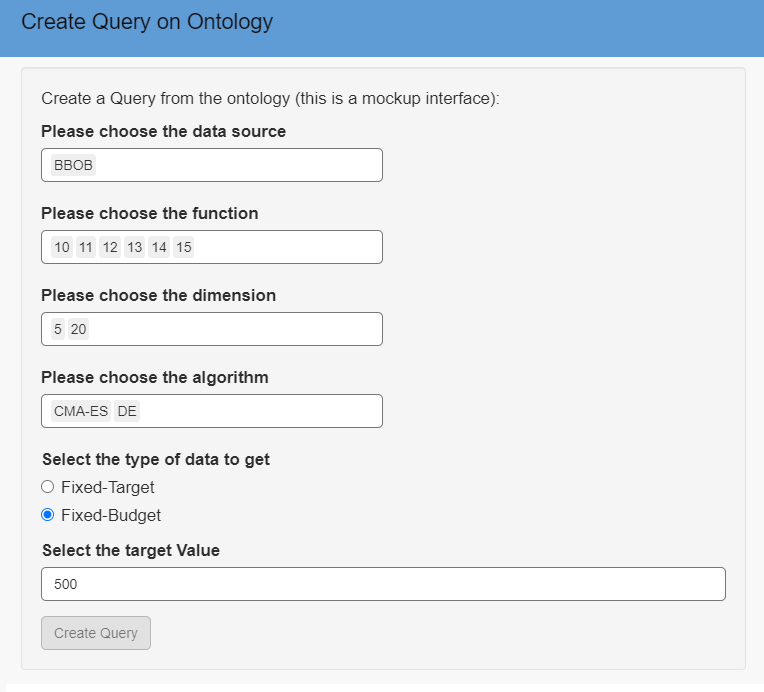}
    \caption{A GUI mockup for querying the knowledge base.}
    \label{fig:mockup}
\end{figure}
A non-exhaustive list of queries that are currently supported by the OPTION knowledge base is as follows: (1) For a given algorithm (e.g., CMA-ES) and fixed budget scenario (e.g., between 1000 - 2000 function evaluations), return the noise-free fitness - Fopt  calculated for a set of benchmark problems (e.g., f1 and f7) and a set of problem instances (e.g., the first 5 instances); (2) For a given algorithm (e.g., CMA-ES) and fixed target, return the number of function evaluations needed to reach the target; (3) For multiple algorithms and fixed budget return the noise-free fitness - Fopt calculated for a set of benchmark problems (e.g., f1-f5); (4) For a set of instances from a given benchmark problem, return all noise-free fitness - Fopt values for all (or a set of) algorithms with a fixed budget of 1000 function evaluations; (5) For a given study, return the provenance information.

SPARQL queries can become very complex and sometimes are seen as a bottleneck to the wider acceptance of Semantic Web technologies. SPARQL query construction is an error-prone and time-consuming task, which requires expert knowledge of the whole stack of semantic technologies. To facilitate the process of querying and to avoid the need of directly writing SPARQL queries, we propose the use of a GUI (for which a mock-up is depicted in Figure~\ref{fig:mockup}) to alleviate the problem explained above. Such a GUI would provide end-users with a search tool through which they can parameterize the predefined queries, without knowing the exact semantic data model used for the annotations. The users would construct the query by providing the search parameters of interest via input fields or predefined drop-down menus that contain ontology terms and the SPARQL queries would be generated in the background.

\section{Conclusions \& Future work}

The development of the OPTION ontology is a big step forward in the process of making benchmark data more reusable and interoperable. By annotating a large subset of the BBOB-data, we have shown the potential of the ontology to aid with data integration, while simultaneously providing powerful querying capabilities for direct analysis of the required datasets. This significantly reduces the time needed to collect data across many functions and algorithms, while being flexible in the type of performance perspective (i.e., fixed-budget, fixed-target) to be used.

Our next steps are as follows. First, we will annotate more data sets, starting with those provided by IOHprofiler~\cite{DoerrYHWSB20,VermettenCASH} and those of Nevergrad~\cite{nevergrad}. Further extensions include a web-based GUI, similar to what is shown in Figure~\ref{fig:mockup}. We believe this to be important for better adoption of our ontology by the optimization heuristics community, whose members are not necessarily familiar with SPARQL queries. This can then be further extended to include direct integration with existing post-processing tools, e.g., IOHanalyzer~\cite{IOHanalyzer}, which would greatly simplify the process of robust algorithm comparison.

\begin{acks}
The authors acknowledge the support of the Slovenian Research Agency through research core grants No. P2-0103 and P2-0098, project grants No. J2-9230 and Z2-1867, and young researcher grant No. PR-09773 to AK, as well as the EC through grant No. 952215 (TAILOR). Our work is also supported by the Paris Ile-de-France region.
\end{acks}